\documentclass[11pt, a4paper, logo, onecolumn, copyright, numbering]{googledeepmind}

\usepackage[authoryear, sort&compress, round]{natbib}
\title{Model Integrity when Unlearning with T2I Diffusion Models}

\correspondingauthor{arischioppa@google.com}

\author[1]{Andrea Schioppa}
\author[1]{Emiel Hoogeboom}
\author[1]{Jonathan Heek}

\affil[1]{Google DeepMind}

\begin{abstract}
The rapid advancement of text-to-image Diffusion Models has led to their widespread public accessibility. However these models, trained on large internet datasets, can sometimes generate undesirable outputs. To mitigate this, approximate Machine Unlearning algorithms have been proposed to modify model weights to reduce the generation of specific types of images, characterized by samples from a ``forget distribution'', while preserving the model's ability to generate other images, characterized by samples from a ``retain distribution''. While these methods aim to minimize the influence of training data in the forget distribution without extensive additional computation, we point out that they can compromise the model's integrity by inadvertently affecting generation for images in the retain distribution. Recognizing the limitations of FID and CLIPScore in capturing these effects, we introduce a novel retention metric that directly assesses the perceptual difference between outputs generated by the original and the unlearned models. We then propose unlearning algorithms that demonstrate superior effectiveness in preserving model integrity compared to existing baselines. Given their straightforward implementation, these algorithms serve as valuable benchmarks for future advancements in approximate Machine Unlearning for Diffusion Models.
\end{abstract}

\usepackage{multirow}
\newcommand\soutpars[1]{\let\helpcmd\sout\parhelp#1\par\relax\relax}
\long\def\parhelp#1\par#2\relax{%
  \helpcmd{#1}\ifx\relax#2\else\par\parhelp#2\relax\fi%
}

\usepackage{xspace} % Optional final space
\def\mixeralgo{Saddle\xspace}
\def\overwritealgo{OVW\xspace}

\newcounter{parcounter}[section]
\renewcommand{\theparcounter}{\arabic{section}.\arabic{parcounter}}
\newcommand\numparagraph[1]{%
    \refstepcounter{parcounter}%
    \paragraph{\S\theparcounter\ #1}
}
\usepackage{amsmath,amsfonts,bm}

\def\normal#1;#2.{\mathcal{N}(#1,#2)}
\def\diffloss{L_{\text{diff}}}
\def\mixerloss{L_{\text{Saddle}}}
\def\helploss{L_{\text{help}}}
\def\ovwloss#1{%
  \def\tempa{}%
  \def\tempb{#1}%
  \ifx\tempa\tempb
    L_{\text{OVW}}
  \else
    L^{#1}_{\text{OVW}}
  \fi
}
\def\consloss{L_{\consmet}}
\def\helploss{L_{\text{help}}}
\def\noisefn#1;#2;#3.{\epsilon_{#1}(#2 | #3)}
\def\consmet{\mathcal{I}}
\def\convsteps{\text{conv}_s}
\def\unlmet{p_{\text{Un}}}
\def\lpipsmet{\text{LPIPS}}
\def\dataa{\mathcal{D}}
\def\datar{\mathcal{D}_r}
\def\datargen{\mathcal{D}^{\text{gen}}_r}
\def\dataf{\mathcal{D}_f}
\def\datafalt{\mathcal{\tilde D}_f}
\def\datah{\mathcal{D}_h}
\def\promptf{\mathcal{P}_f}
\def\promptr{\mathcal{P}_r}
\def\prompth{\mathcal{P}_h}
\DeclareFixedFont{\ttb}{T1}{txtt}{bx}{n}{10} % for bold
\DeclareFixedFont{\ttm}{T1}{txtt}{m}{n}{10}  % for normal

\usepackage{color}

\definecolor{deepblue}{rgb}{0,0,0.5}
\definecolor{deepred}{rgb}{0.8,0,0}
\definecolor{deepgreen}{rgb}{0,0.5,0}
\definecolor{grey1}{rgb}{0.5,0.5,0.5}
\usepackage{listings}

\newcommand\pythonstyle{\lstset{
		language=Python,
		basicstyle=\small\ttm,
		otherkeywords={self},             % Add keywords here
		keywordstyle=\small\ttb\color{deepblue},
		emph={},          % Custom highlighting
		emphstyle=\small\color{deepred},    % Custom highlighting style
		stringstyle=\small\color{deepgreen},
		commentstyle=\small\color{grey1}\ttm,
		frame=tb,                         % Any extra options here
		showstringspaces=false,            % 
		breaklines=true,
		tabsize=2
}}
\lstnewenvironment{python}[1][] {
	\pythonstyle
	\lstset{#1}
}{}

\newcommand\pythoninline[1]{{\pythonstyle\lstinline!#1!}}

\begin{document}

\maketitle

% Prevent Main sections to show up in the Appendix
\addtocontents{toc}{\protect\setcounter{tocdepth}{0}}
\section{Introduction}

\paragraph{Context}
Recent years have witnessed rapid advancements in text-to-image generative models, particularly Diffusion Models, leading to the widespread availability of powerful models like Stable Diffusion and Midjourney \citep{Rombach2021HighResolutionIS, Midjourney2023}. However, these models may inadvertently learn undesirable concepts from their training data, posing challenges at deployment \citep{fan2024salun, heng2023selective, wu2024erasediff}. Consequently, research into methods for removing such concepts from model weights has gained significant traction. One promising avenue is Machine Unlearning (MU), which broadly aims to mitigate the influence of specific training data subsets (forming the ``forget set'' $\dataf$) after pre-training \citep{Shaik2023exploringunl}. Due to the substantial computational costs associated with pre-training large models, both retraining and exact MU approaches \citep{Guo2019CertifiedDR, thudi22necessity, Bourtoule2019unlearning} prove impractical. This has spurred the development of more feasible approximate MU techniques \citep{Shaik2023exploringunl, izzo21approximatedeletion, fan2024salun}, often referred to as ``forgetting'' or ``erasing concepts'' in the context of Diffusion Models \citep{Gandikota_2023_ICCV, heng2023selective}.

\paragraph{Motivation: Ensuring Model Integrity After Unlearning}
Evaluating the effectiveness of approximate MU algorithms necessitates assessing both the extent of forgetting and the preservation of model quality/integrity on the retain dataset
$\datar$. Existing approaches for diffusion models primarily rely on FID \citep{Heusel2017FID} and CLIPScore \citep{hessel21clipscore} metrics, applied to a set of retain prompts. However, recent work \citep{Zhang2024UnlearnCanvasAS} has criticized this methodology, advocating for the development of new benchmark datasets and the use of fine-tuned classifiers to gauge in-domain and cross-domain retention
because quantifying retention using a single metric lacks precision in the generative context~\citep[Sec.~2, pg.~4]{Zhang2024UnlearnCanvasAS}. While acknowledging the potential benefits of such an approach
(e.g.~precision and granularity on sub-domains of the retain dataset), we highlight several limitations. Firstly, developing new benchmarks or fine-tuning classifiers incurs significant costs. Additionally, using multiple fine-tuned classifiers results in a non-well-ordered set of in-domain and out-of-domain scores, complicating evaluation. Furthermore, classifiers may not adequately capture subtle qualitative changes in model outputs after unlearning (Fig.~\ref{fig:qual_eval}, \S\ref{par:qualitative}). Similarly, FID and CLIPScore may fail to detect unintended changes on specific subsets of the retain set, as illustrated in Fig.~\ref{fig:fid_inadequate}, where forgetting Van Gogh's style inadvertently impacts the generation of Vermeer's style. While FID could be computed on subsets of the retain set, this is computationally expensive due to its distributional nature and potential scaling issues across subsets (see, for example, \citep[Tab.~1]{rule_them_all2024} where FID ranges can change significantly across subsets). To address these challenges, we propose a novel integrity metric (Sec.~\ref{sec:methodology}) that directly quantifies the similarity between images generated by the unlearned and original checkpoints on the retain set. This approach by-passes the need for ad-hoc evaluation datasets or fine-tuned classifiers, offering a more practical alternative to estimating model closeness via parameter posteriors or model predictions~\citep{hu2022membershipinferenceattacksmachine}. Our empirical analysis reveals that current methods often struggle to maintain integrity, leading to unintended consequences such as the removal of multiple artistic styles when targeting a single style. This underscores a gap in existing approaches, motivating our development of two novel algorithms that prioritize integrity preservation. 

\paragraph{Contributions}
We summarize our contributions as follows:
\begin{enumerate}
\item Integrity Metric: We introduce the integrity metric, $\consmet$ (\S\ref{par:cons_met}), providing a quantitative measure to assess preservation of generated outputs after unlearning.
\item Integrity-Driven Algorithms: We present two novel algorithms designed to prioritize integrity preservation, addressing both supervised and unsupervised scenarios.
\item Empirical validation: Our proposed algorithms demonstrate:
\begin{itemize}
\item Improved integrity and reduced side effects: Superior performance in maintaining integrity and minimizing unintended consequences (Table~\ref{tab:quant_comp}, Figs.~\ref{fig:fid_inadequate}, \ref{fig:qual_eval}).
\item Simplicity: straightforward implementation, facilitating future research by serving as effective baselines (implementation details available in the Appendix~\ref{sec:algo_implementation}).
\item Better behavior in data-limited scenarios: they outperform current methods when one needs to rely on generated images as the original training data is inaccessible (\S\ref{par:abl_pre_data}).
\end{itemize}
\end{enumerate}

\paragraph{Reproducibility} We include Python code snippets of the implementation in Appendix~\ref{sec:algo_implementation}.

\section{Related Work}
\label{sec:related_word}

\paragraph{Machine Unlearning}
In the context of Machine Unlearning (MU), one considers two datasets: $\dataf$ (forget set) and $\datar$ (retain set). An initial model checkpoint, $\theta_0$, is trained on the combined data $\dataf \cup \datar$. The objective of MU is to modify $\theta_0$ to approximate a checkpoint trained solely on $\datar$. While ideally, the posterior distribution of weights after MU should closely resemble that of a model trained exclusively on $\datar$, verifying this at the scale of text-to-image Diffusion Models is impractical. Thus, practical approximate MU methods focus on preventing the model from generating images in $\dataf$ when presented with their corresponding prompts, $\promptf$.  We broadly classify these methods into two categories: those that do not allow specifying a target distribution for prompts in $\promptf$ (unsupervised, \S\ref{par:unsup_meth}) and those that do (supervised, \S\ref{par:sup_meth}).

\numparagraph{Unsupervised methods}\label{par:unsup_meth}
Unsupervised methods aim to modify the predicted noise distribution or increase the loss on the forget dataset ($\dataf$) without specifying a target image distributions for its prompts $\promptf$.  NegGrad~\citep{Golatkar_2020_CVPR} directly performs gradient ascent on the training loss associated with $\dataf$, while ESD~\citep{Gandikota_2023_ICCV} reverses the sign of the guidance component in the noise on $\dataf$. A key limitation of both NegGrad and ESD is the absence of a retention objective to preserve performance on the retain dataset ($\datar$). EraseDiff~\citep{wu2024erasediff} addresses this shortcoming by fine-tuning on $\datar$; on $\dataf$
it replaces the Gaussian noise distribution with a uniform one.

\numparagraph{Supervised methods}\label{par:sup_meth}
Supervised methods replace the target image distributions for prompts in $\dataf$, creating a new target distribution $\datafalt$ (e.g., in the case of MNIST one might remap digit 1 to a dark background if the forget set is digit 1). Selective Amnesia~\citep{heng2023selective} (SA) achieves this by fine-tuning the model on the combined dataset  $\datafalt\cup\datar$ and employing the Fisher Information Matrix (FIM) to keep unlearned parameters ($\theta_u$) close to their initial values ($\theta_0$), leveraging second-order information to influence the more ``salient'' parameters. Our work demonstrates that SA can still impact data in $\datar$ that exhibits proximity to data in $\dataf$. A compelling advantage of supervised methods is explicit control over the target distribution on $\dataf$ (see \S\ref{par:qualitative} for qualitative results). The recent SalUn method~\citep{fan2024salun} proposes pre-computing a saliency mask to target only the most relevant parameters during unlearning, offering a computationally efficient alternative to FIM approximation. While masks can be computed on any loss, we utilize it on top of ESD, following the code release of~\citep{fan2024salun}.

\paragraph{Robustness, adversarial and text-based methods}
\citep{pham2024circumventing} demonstrate that current MU methods are vulnerable to text-inversion attacks and can be bypassed if the user has access to text embeddings.  \citep{xiong2024editing} propose a scalable model-editing technique on the text encoder to control generation in diffusion models, a method that can be used for MU. However, our work differs crucially in that we assume a fixed, non-updatable text-encoder (e.g., a pre-trained LLM with pre-computed embeddings on the pre-training data). This limitation aligns with real-world situations where the text encoder might be a large, pre-trained language model that is kept frozen, particularly if the original text embeddings have already been pre-computed on the pre-training data. \citep{zhang2024defensive} enhance the robustness of ESD when applied to Stable Diffusion (SD) by incorporating adversarial training. To mitigate the FID decrease associated with adversarial training, they introduce a loss term similar to our integrity term, $\consloss$. However, our work diverges in several key aspects: 1) We motivate $\consloss$ from a integrity preservation perspective, benchmark it against using $\diffloss$ on the retain set $\datar$, and demonstrate that $\consloss$ alone is insufficient for retention when used with Supervised Methods. Additionally, we operate under the assumption that the text encoder cannot be modified, thus precluding the use of those adversarial attacks. \citep{getwhatyouwant2024} investigates the impact of the [EOT] token(s) in the CLIP encoder within Stable Diffusion and subsequently proposes a content suppression mechanism based on text embedding manipulations. In our setup, the [EOT] token is absent as the output corresponding to padding tokens is masked in the encoder. \citep{conceptpresgradients2024} employs an adversarial network to design a forget loss on the forget set $\dataf$ and utilizes gradient surgery between the retain $\datar$ and forget $\dataf$ sets to enhance retention. We bypass the need for gradient surgery by introducing a help set of prompts in the supervised methods scenario. The construction of this help set is straightforward (see \S\ref{par:construction_help_prompts}) and doesn't necessitate estimating distances between concepts in $\datar$ and $\dataf$.

\paragraph{Sparser or Structured updates}
Our study of MU focuses on modifying the original model weights without resorting to sparse or structured updates. For sparse updates we mean updates that
modify a smaller subset of model parameters, say the cross-attention layers. On the other hand, structured updates assume a lower rank structure
on the update and can modify either all the model parameters or a subset of them. A key benefit of these approaches is to reduce the memory footprint for large models
allowing implementation without needing to partition large models across accelerators. Other
research streams have explored such updates; for example, \citep{forgetmenot2024} updates the cross-attention in the U-NET to mask concepts in Stable Diffusion. \citep{Gandikota_2023_ICCV} also experimented with limiting updates solely to the cross-attention. In our early experiments with the UViT, we observed that updating only the cross-attention led to either worse FID compared to updating the entire network or a worse unlearning score $\unlmet$. \citep{rule_them_all2024} avoids modifying model weights and instead uses a rank-one adapter to control the model. To address interference between erasure and retaining existing concepts, they propose up-sampling data from $\datar$ based on an importance score derived from CLIP embeddings. By contrast, our focus lies in modifying the original weights (using model parallelism), and we demonstrate that high retention can be achieved by acting solely on the original weights, without the need for an additional network. Furthermore, we present a simpler approach to retention in the supervised setup by creating and filtering a set of prompts using a sentence similarity model (see \S\ref{par:construction_help_prompts}). Finally, early experiments with LORA updates on the model weights showed less competitive performance compared to full updates: this doesn't contradict the findings of~\citep{rule_them_all2024} as they utilize LORA to construct an additional network that conditions the model based on the input concept, whereas we explored LORA for updating the original weights.

\section{Methodology}
\label{sec:methodology}

\paragraph{Text-to-Image Diffusion Models}
The diffusion process transforms an image $x$ into a sample drawn from the standard normal distribution $\normal 0;1.$;
a diffusion model parameterized by $\theta$ gives a noise estimator $\noisefn\theta;\cdot;c.$ (where $c$ is the conditioning
text input) such that if $x_t$ is the noised image at time step $t$, $\noisefn\theta;x_t;c.$ is trained to minimize the
de-noising error objective:

\begin{equation}
    \diffloss(\theta) = E_{(x,c)\sim\dataa, t\sim [0,T], \varepsilon\sim\normal 0;1.} \|\varepsilon -  \noisefn\theta;x_t;c.\|_2^2.
\end{equation}
Note that $\diffloss$ can be formulated in a more general form that involves a time-step dependent weight $w(t)$ (e.g.~computable from the signal to noise ratio, see~\cite{vdm_kingma21}).
In our work we set $w(t)=1$ for consistency with previous work in the unlearning diffusion literature, but our methods are compatible with losses that give a different
weight to each time steps. 
\numparagraph{Integrity Metric}\label{par:cons_met}
\begin{figure}
    \centering
    \begin{tabular}{c|ccccccc}
    
    & \scriptsize{Base}
    & \scriptsize{EraseDiff}
    & \scriptsize{SA}
    & \scriptsize{ESD}
    & \scriptsize{\mixeralgo (ours)}
    & \scriptsize{\overwritealgo (ours)}
    & \scriptsize{\overwritealgo $-\helploss$}
    \\
    \raisebox{0.05\textwidth}{Forget}
    &   \includegraphics[width=0.1\textwidth]{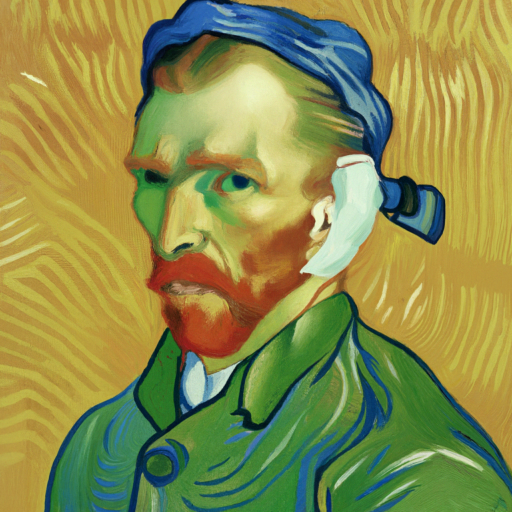}
    &   \includegraphics[width=0.1\textwidth]{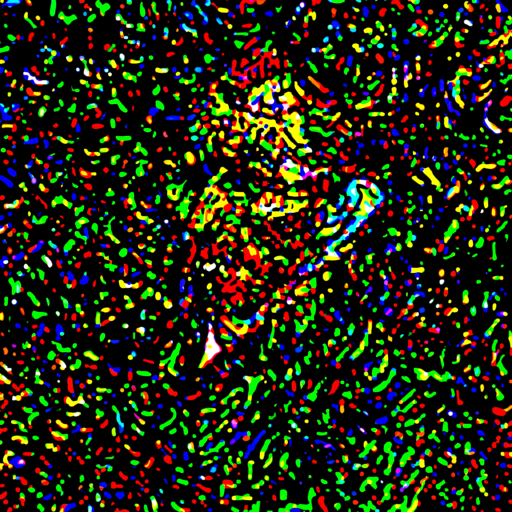}
    &   \includegraphics[width=0.1\textwidth]{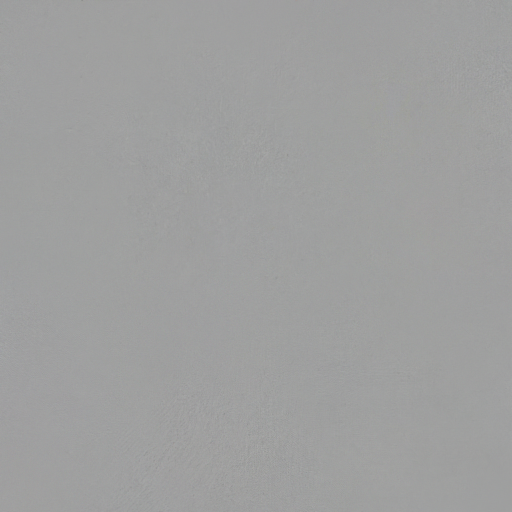}
    &   \includegraphics[width=0.1\textwidth]{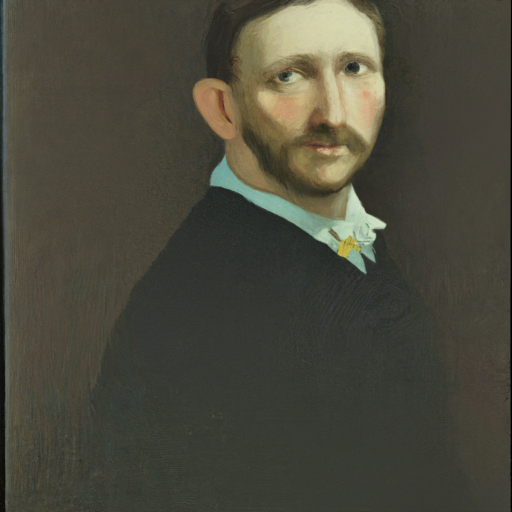}
    &   \includegraphics[width=0.1\textwidth]{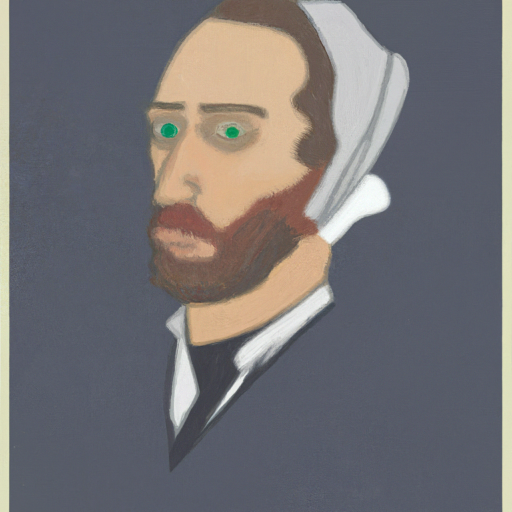}
    &   \includegraphics[width=0.1\textwidth]{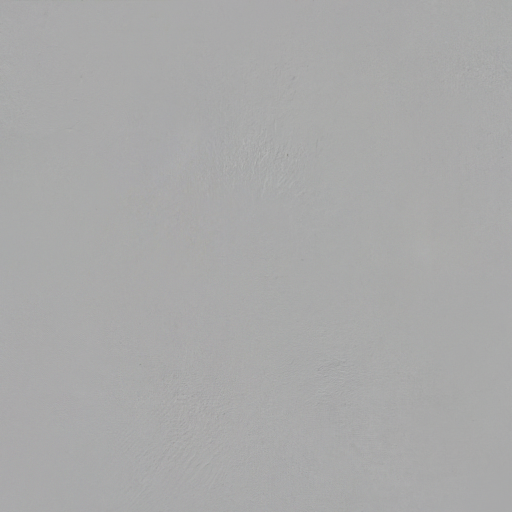}
    & \includegraphics[width=0.1\textwidth]{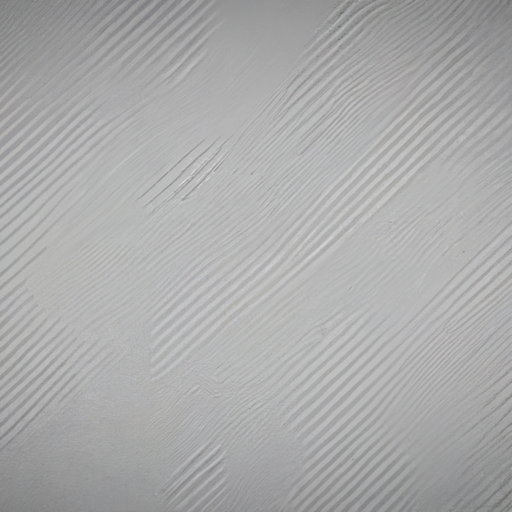}
    \\
    \midrule
    \raisebox{0.05\textwidth}{Retain}
    &   \includegraphics[width=0.1\textwidth]{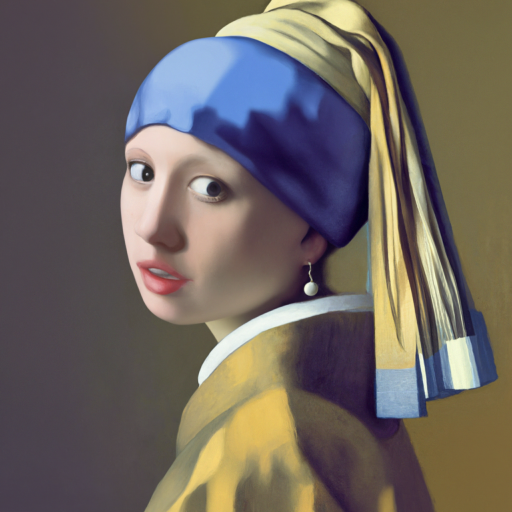}
    &   \includegraphics[width=0.1\textwidth]{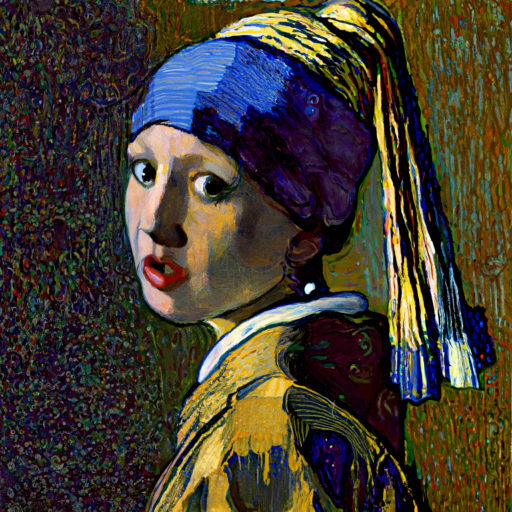}
    &   \includegraphics[width=0.1\textwidth]{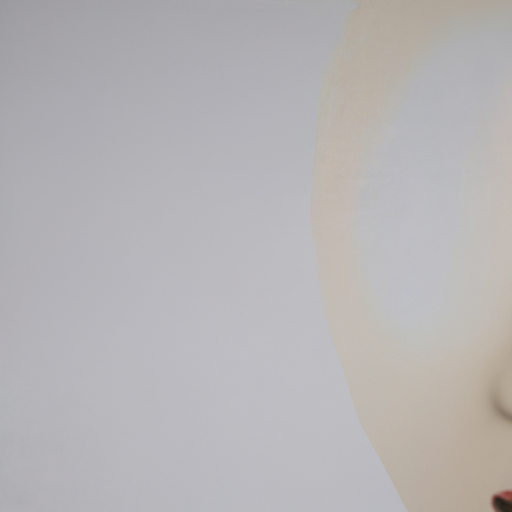}
    &   \includegraphics[width=0.1\textwidth]{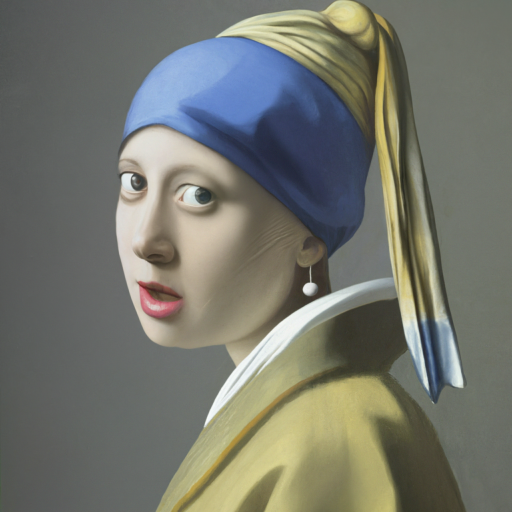}
    &   \includegraphics[width=0.1\textwidth]{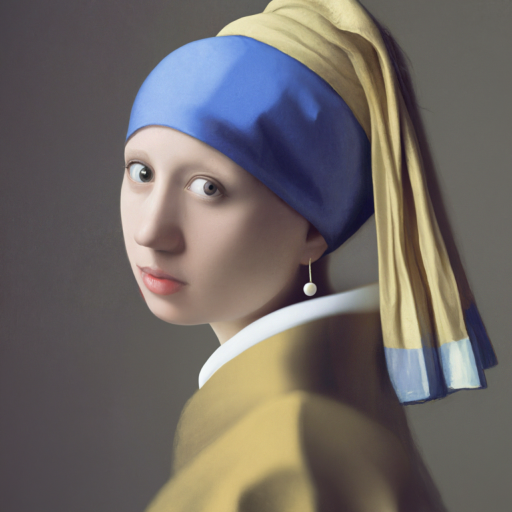}
    &   \includegraphics[width=0.1\textwidth]{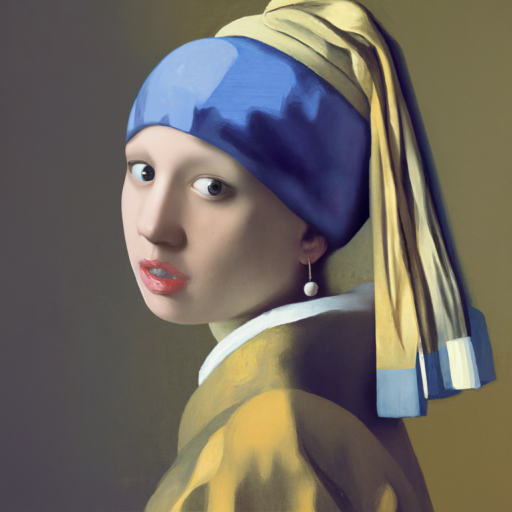}
    &  \includegraphics[width=0.1\textwidth]{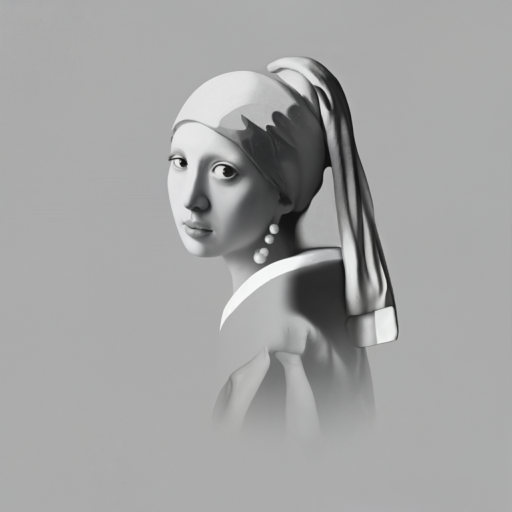}
    \\
    \midrule
    $\lpipsmet$ & 0.0 & 0.43 & 0.67 & 0.32 & 0.17 & 0.04 & 0.58
    \end{tabular}
    \caption{Forgetting Van Gogh affects generation on a different painter, Vermeer. Except for ESD, all models have comparable FID (within $0.5$) and CLIPScore (within $0.2$), 
    see Table~\ref{tab:quant_comp}, so this
    effect is not captured by FID or CLIPScore. Our methods preserve generation on Vermeer while forget Van Gogh. Removing the help loss terms $\helploss$
    from \overwritealgo interferes with image generation on Vermeer. Generated images share the random seed on each row. $\lpipsmet$ is computed wrt.~the reference image generated by the base
    checkpoint.
    }
    \label{fig:fid_inadequate}
\end{figure}
In Figure~\ref{fig:fid_inadequate}, we illustrate a scenario where the dataset $\dataf$ comprises paintings in Van Gogh's style, while the retain set $\datar$ contains everything else, including works by Vermeer. Despite achieving FID@30k and CLIPScore comparable to the base checkpoint, both EraseDiff and SA exhibit significant forgetting of Veermer and are negatively impacted in their ability to generate images in Vermeer's style. This underscores the limitations of FID@30k and CLIPScore in capturing the retention of generation quality on the retain set. One potential approach is to create multiple evaluation subsets from $\datar$ and report FID/CLIPScore on each. For example,
some subsets might represent close concepts, e.g. painting styles: the painting style of Monet, Michelangelo, \dots\ Other subsets might correspond to more distant
concepts like furniture objects, animals, celebrities \dots\ However, this approach suffers from several drawbacks; first, the effort required to construct these subsets;
second, challenges in interpreting FID metrics that are potentially scaled differently across subsets (compare \citep[Tab.~1]{rule_them_all2024}); third, the inability to establish a total order on models due to varying performance across subsets (tuples of scores on such subsets are partially but not well-ordered).
We propose a simpler alternative: the integrity metric $\consmet$: $\consmet$ directly compares the original checkpoint $\theta_0$ and the unlearned checkpoint $\theta_u$
on a distribution of retain prompts $\promptr$. It quantifies the difference in images generated for these prompts between the two checkpoints using the $\lpipsmet$ metric~\citep{lpips2018}:

\begin{equation}
\consmet(\theta_0, \theta_u, \promptr) = 
E_{c\sim \promptr, g\sim S} \lpipsmet(G(\theta_0, c, g), G(\theta_u, c, g)),
\end{equation}
where $G$ is the image generating function, $c$ is a text prompt, and $g$ is a random seed. Formally, the set of retain $\promptr$ consists of the prompts in $\datar$ (whose elements are pairs of prompts and images);
however, from a practical perspective, it can be taken as a set of prompts corresponding to concepts from $\datar$ seen during training.
While we utilize $\lpipsmet$ (scaled between 0 and 1), other metrics like the pixel-level $l_1/l_2$ distances could be employed. To give an intuition regarding the definition of $\lpipsmet$, it is a perceptual distance that measures the distance between images using features extracted by a neural network instead than using
pixel-level features; this approach has several advantages pointed out by~\citep{lpips2018}. It's important to note that $\consmet$ depends on the choice of retain prompts in $\promptr$. Since using all prompts from 
$\datar$ is often infeasible, one can sample a subset. In the context of Figure~\ref{fig:fid_inadequate}, $\consmet$ is significantly higher for EraseDiff and SA compared to our proposed methods, both when analyzing a specific image (Figure~\ref{fig:fid_inadequate}) and over a set of retain prompts (Table~\ref{tab:quant_comp}).

We chose the term \emph{integrity} because \emph{retention} is typically used in the unlearning literature to describe preserving the value of an evaluation metric (e.g., accuracy for classifiers) on
$\datar$. As we demonstrated, retaining FID or CLIPScore on $\datar$ is insufficient. Our metric measures consistency with the original checkpoint $\theta_0$ when generating images. We avoided \emph{consistency} as it is already used in the diffusion literature for distillation; \emph{integrity} conveys the intention of preserving the behavior of $\theta_0$ on $\datar$.

\numparagraph{Unlearning Algorithms}\label{par:unlearn_algos}
We now focus on designing unlearning algorithms that take $\consmet$ into account. While directly optimizing $\consmet$ is possible, it's computationally expensive due to the multiple sampling steps required for image generation and the back-propagation through the $\lpipsmet$ neural network. To circumvent this, we leverage two observations:
\begin{itemize}
\item Images that are close in pixel space are also close in $\lpipsmet$.
\item 
If the noise distributions $\noisefn\theta_u;x_t;c.$ and $\noisefn\theta_0;x_t;c.$ are close in $l_2$, the generated images $G(\theta_0, c, g)$ and $G(\theta_u, c, g)$ will be close in pixel space.
\end{itemize}
Based on these observations, we define an integrity loss:

\begin{equation}\label{eq:consloss}
\consloss(\theta) = \beta \cdot E_{c\sim\promptr, t\sim[0,T], x\sim\datar(c)} \|( \noisefn\theta;x_t;c. - \noisefn\theta_0;x_t;c.\|^2,
\end{equation}
where $\beta$ is a boosting hyper-parameter and where one might introduce a time-step dependent weight $w(t)$ (compare~\cite{vdm_kingma21}). This loss avoids the need for multiple sampling steps as we don't generate images directly. In Equation \ref{eq:consloss}, we need to specify how to sample the image $x$ from $\datar$.
While most previous work focused on Stable Diffusion where $\datar$ is inaccessible, our case allows access to $\datar$ as we pre-trained our model. If available, we can sample $x$ corresponding to the prompt $c$ (denoted as $x\sim\datar(c)$). If $\datar$ is inaccessible, we can generate $x$ using the base checkpoint, potentially pre-generating a dataset of images for retention as in~\citep{heng2023selective}. In \S\ref{par:abl_pre_data}, we provide ablations showing that $\consloss$
works well both with and without access to the pre-train data. We also demonstrate that the alternative approach of using $\diffloss$
restricted to $\datar$ leads to significant degradation in FID.

We propose our first algorithm, \mixeralgo, which operates without supervising the distribution on the forget set of prompts. It performs gradient descent on the objective:

\begin{equation}\label{eq:mixerloss}
\mixerloss(\theta) = -E_{(x,c)\sim\dataf, t\sim [0,T], \varepsilon\sim\normal 0;1.} \|\varepsilon -  \noisefn\theta;x_t;c.\|_2^2 + \consloss(\theta),
\end{equation}
as the first term is the opposite of $\diffloss$ on the forget set, \mixeralgo seeks a saddle point of $\mixerloss$, hence the choice for the name. 

For algorithms that supervise the target distribution, we start with SA \citep{heng2023selective}. Early experiments showed that we can drop the expensive Fisher-Information term in SA. If $\datafalt$
is the distribution SA replaces $\dataf$ with (see \S\ref{par:sup_meth}), one might optimize:
\begin{equation}\label{eq:owv_part_1}
\ovwloss{1} = E_{(x,c)\sim\datafalt, t\sim [0,T], \varepsilon\sim\normal 0;1.} \|\varepsilon -  \noisefn\theta;x_t;c.\|_2^2 + \consloss(\theta).
\end{equation}
However, in Section \ref{par:abl_help}, we show that this still leads to overwriting related concepts (Figure~\ref{fig:fid_inadequate}).
To address this, we construct a small subset of help prompts $\prompth$ related to $\dataf$ (details in \S\ref{par:construction_help_prompts}). We generate images from 
$\prompth$ using the base checkpoint $\theta_0$, obtaining a dataset $\datah$.
We define $\helploss$ similarly to $\consloss$ but sample $(x,c)$ from $\datah$. 
Then, we define:
\begin{equation}\label{eq:owv_part_2}
\ovwloss{2} = E_{(x,c)\sim\datafalt, t\sim [0,T], \varepsilon\sim\normal 0;1.} \|\varepsilon -  \noisefn\theta;x_t;c.\|_2^2 + \helploss(\theta),
\end{equation}
and propose the \overwritealgo that alternates a gradient descent step on $\ovwloss{1}$ with one on $\ovwloss{2}$. The name \overwritealgo is an acronym of
Ovewrite.

%% TODO: FIND PROPER PLACE
\begin{table*}[t]
\caption{Quantitative comparison of methods on image quality (FID), following text instructions (CLIPScore), consistency with the
original checkpoint ($\consmet$) and probability that generated imaged are classified as belonging to $\dataf$ ($\unlmet$). 
$\convsteps$ denotes the number of steps to convergence.
}
\label{tab:quant_comp}
\begin{center}
\begin{tabular}{c|c|ccccc}
\hline
Method & $\dataf$ & FID@30k $\downarrow$ & CLIPScore $\uparrow$ & $\consmet$ $\downarrow$ (\S\ref{par:cons_met}) & $\unlmet$ (\%) $\downarrow$ & $\convsteps$ (k) \\
\hline
(Base model $\theta_0$) & - & 9.21 & 29.5 & 0 & 100 & - \\
\hline
\multirow{3}{*}{NegGrad} & Celebrity & 10.8 & 29.7 &	0.22 & 1.56 & 0.5 \\
 & Artist & 23.3 & 29.4 & 0.32 & 0.06 & 0.3 \\
 & Animal & 101.0 &	25.7 &	0.54 & 48.88 & 0.6 \\
\hline
\multirow{3}{*}{ESD} & Celebrity & 12.1 & 29.7 &	0.21 & 0.78 & 0.6 \\
& Artist & 14.2 & 28.1 &	0.33 & 0.90 & 1 \\
& Animal & 23.1 & 25.0 &	0.45 & 47.57 & 8 \\
\hline
\multirow{3}{*}{SalUn} & Celebrity & 11.2 &	29.7 & 0.22 & 2.34 & 0.4 \\
& Artist & 13.2 & 28.3 & 0.32 &	 0.89 & 0.8 \\
& Animal & 24.0 & 24.9 & 0.45 &	47.27 & 8 \\
\hline
\multirow{3}{*}{EraseDiff} & Celebrity & 8.7 & 29.6 & 0.22 & 3.90 & 3.0 \\
& Artist & 9.0 & 29.6 &	0.21 & $<10^{-2}$ & 2 \\
& Animal & 8.3 & 29.6 & 0.26 & 88.24 & 6 \\
\hline
\multirow{3}{*}{SA} & Celebrity & 10.8 & 29.5 &	0.16 & 5.47 & 0.9 \\
& Artist & 9.7 & 29.3 & 0.20 & $<10^{-2}$ & 10 \\
& Animal & 20.8 &  28.3 & 0.33 & 0.25 & 20 \\
\hline
\multirow{3}{*}{\mixeralgo \scriptsize{(ours)}} & Celebrity & 9.3 & 29.5	&  0.14 & 2.34 & 0.3 \\
& Artist & 9.4 & 29.4 &	0.09 & 2.47 & 0.4 \\
& Animal & 11.3 &  29.3 & 0.13 & 0.64 & 1.1 \\
\hline
\multirow{3}{*}{\overwritealgo \scriptsize{(ours)}} & Celebrity & 9.1 &	29.6 &	0.08 & 10.9 & 0.8 \\
& Artist & 9.0 & 29.6 &	0.08 & $<10^{-2}$ & 10 \\
& Animal & 11.3 & 29.32 & 0.12 & 9.1 & 20 \\
\hline
\end{tabular}
\end{center}
\end{table*}
\begin{figure}
    \centering
    \begin{tabular}{c|ccccccc}
    & \scriptsize{Base}
    & \scriptsize{EraseDiff}
    & \scriptsize{SA}
    & \scriptsize{ESD}
    & \scriptsize{SalUn}
    & \scriptsize{\mixeralgo (ours)}
    & \scriptsize{\overwritealgo (ours)}
    \\
    \raisebox{0.05\textwidth}{Forget}
    &   \includegraphics[width=0.1\textwidth]{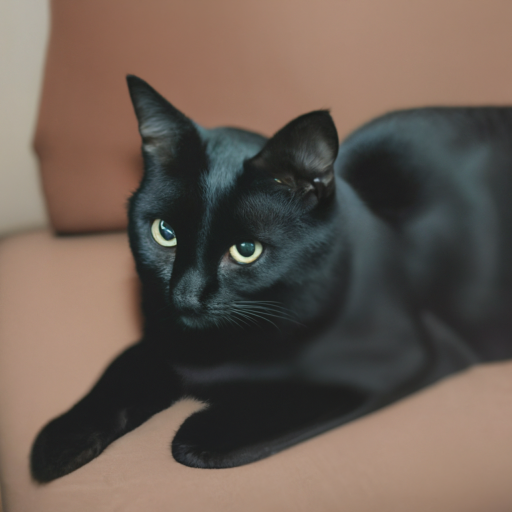}
    &   \includegraphics[width=0.1\textwidth]{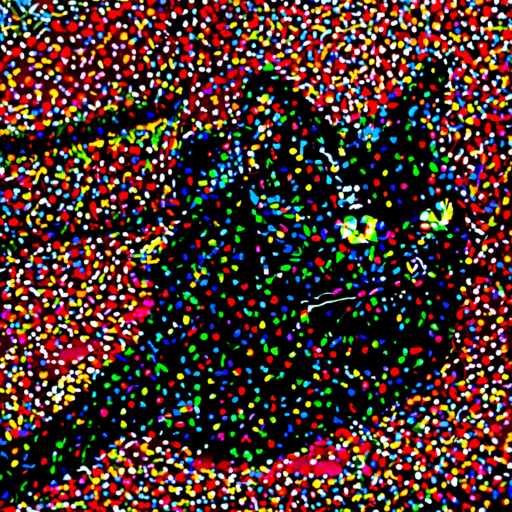}
    &   \includegraphics[width=0.1\textwidth]{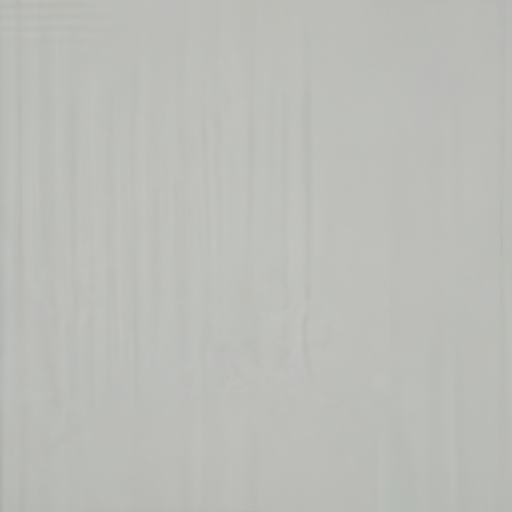}
    &   \includegraphics[width=0.1\textwidth]{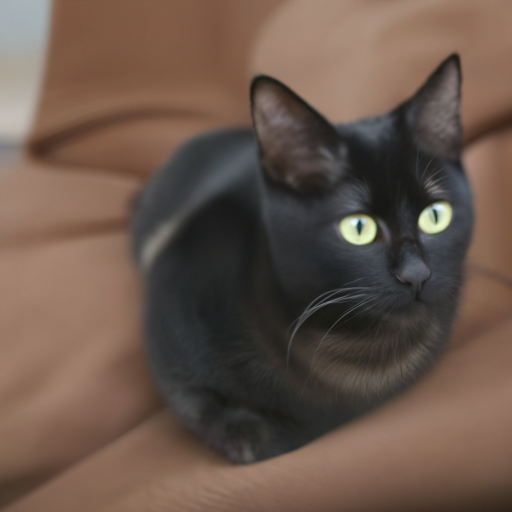}
    &  \includegraphics[width=0.1\textwidth]{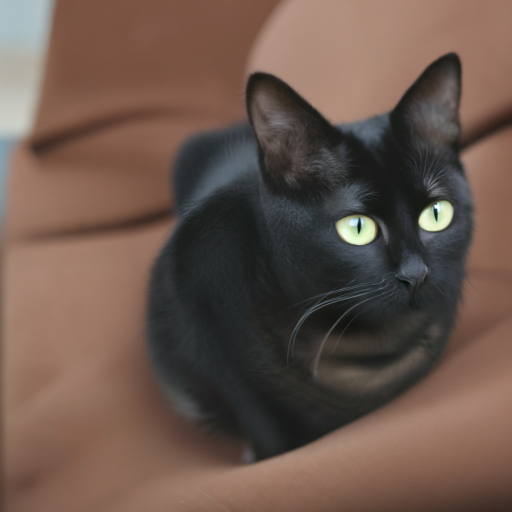}
    &   \includegraphics[width=0.1\textwidth]{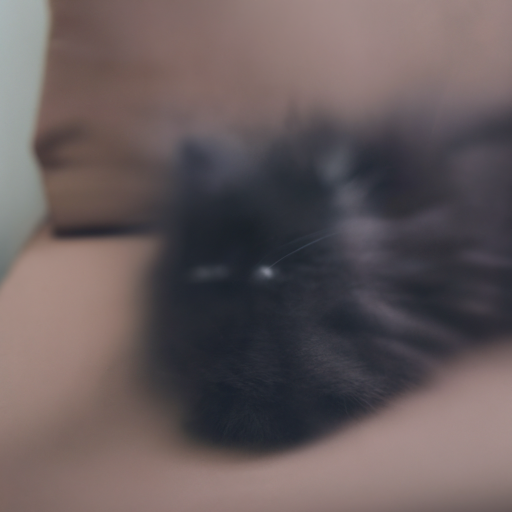}
    &   \includegraphics[width=0.1\textwidth]{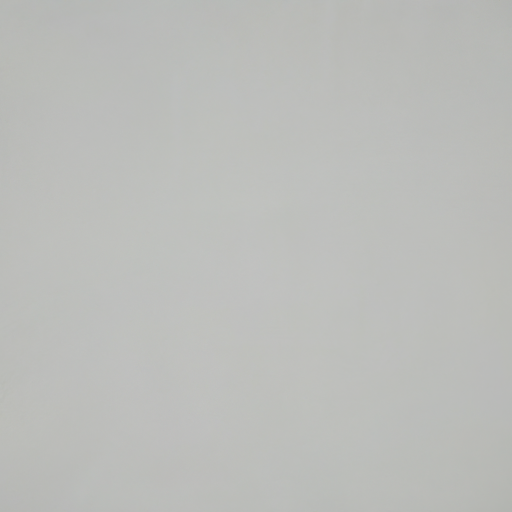}
     \\
    \midrule
    \raisebox{0.05\textwidth}{Retain}
    &  \includegraphics[width=0.1\textwidth]{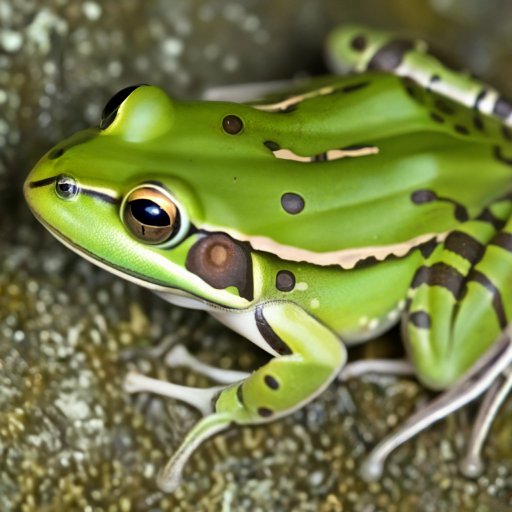}
    & \includegraphics[width=0.1\textwidth]{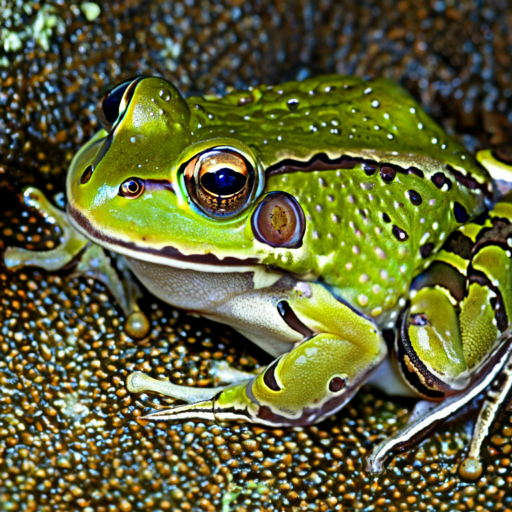}
    & \includegraphics[width=0.1\textwidth]{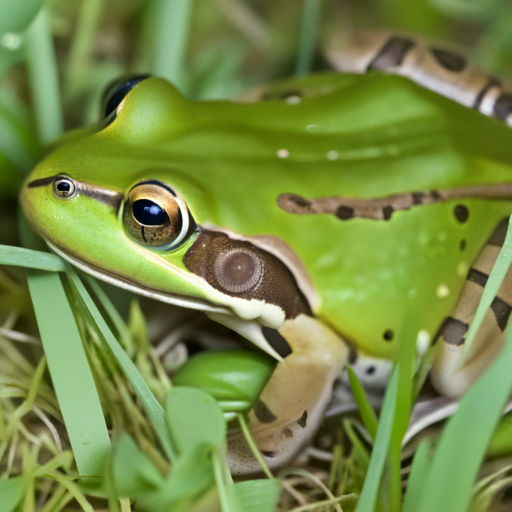}
    & \includegraphics[width=0.1\textwidth]{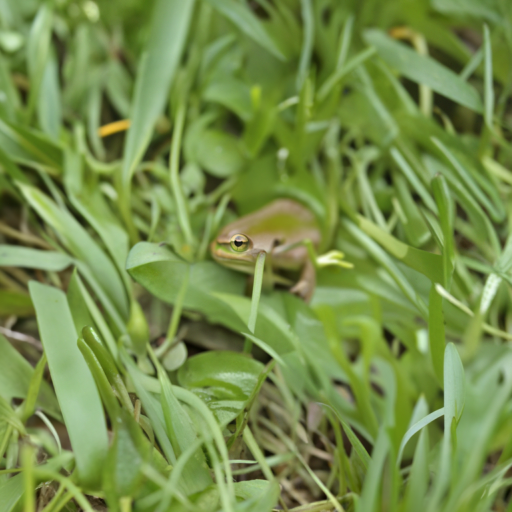}
    &  \includegraphics[width=0.1\textwidth]{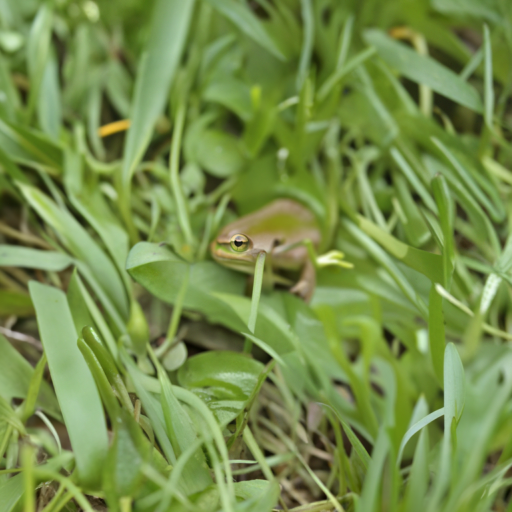}
    & \includegraphics[width=0.1\textwidth]{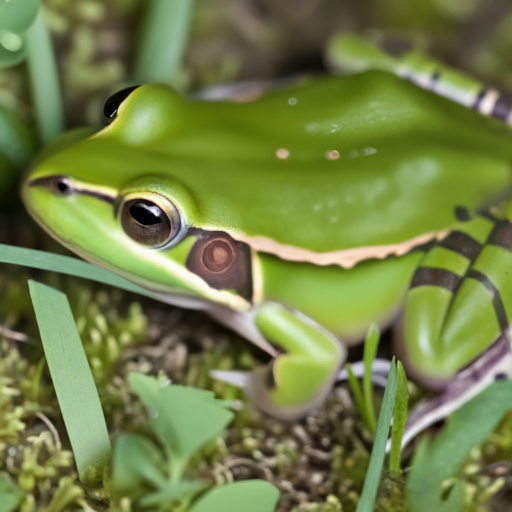}
    & \includegraphics[width=0.1\textwidth]{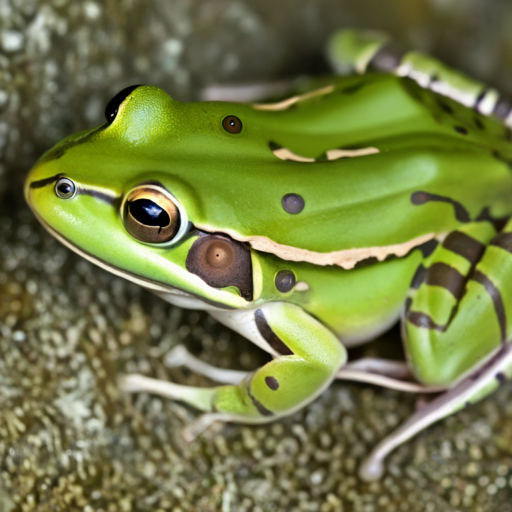}
    \\
    \midrule
    $\lpipsmet$ & 0.0 & 0.45 & 0.40 & 0.59 & 0.59 & 0.32 & 0.08
    \end{tabular}
    \caption{Qualitative comparison of forgetting (or not) cat across methods. Our methods keep generation closest to the original checkpoint on
    the frog. SALUN and ESD produce qualitatively very similar results. For each row the same random seed is used across images.
    $\lpipsmet$ is computed wrt.~the reference image generated by the base
    checkpoint. In the case of~\mixeralgo, $\lpipsmet$ is affected by the appearance of grass under the frog, compare also SA.
    }
    \label{fig:qual_eval}
\end{figure}
\section{Experiments} 
\paragraph{Pre-trained model} We pretrain a text-to-image Diffusion Model following the setup from ``Simple Diffusion'' \citep[Sec.~B]{Hoogeboom2023simplediffusion}.
As text encoder we use T5-XXL~\citep{t5paper} (4B parameters); for the decoder we use standard UViTs as in~\cite[Sec.~3.5]{Hoogeboom2023simplediffusion},
and employ a total number of 2B parameters.
The model is pretrained for
1M steps with a resolution of 128x128 and a batch size of 2048. Text encodings were pre-computed on the whole pre-train dataset. We emphasize that our model is trained
in pixel space and not in latent space as in the case of Stable Diffusion, compare~\citep{Rombach2021HighResolutionIS}.

\paragraph{Tasks}
We evaluate unlearning on three tasks explored in previous work: removing a concept category (specifically, the ``cat'' category,
referred to as \emph{Animal} task)~\citep{fan2024salun}, an artistic style~\citep{Gandikota_2023_ICCV} (we choose Van Gogh, referred to as \emph{Artist} task), and a celebrity~\citep{heng2023selective}
(referred to as \emph{Celebrity} task). These tasks were chosen not only due to their relevance in prior research, but also because they represent forgetting datasets ($\dataf$) of decreasing sizes.  To ensure fair comparison across unlearning methods, we use a consistent learning rate of $5\times 10^{-7}$ for most methods, enabling comparable gradient updates; ESD required a larger learning rate of
$2\times 10^{-5}$ for effective unlearning.  We monitor unlearning progress by inspecting generated images for a set of control prompts every 100 steps
and deem convergence achieved when these images stabilize.  Notably, for the Animal task, even after 20,000 steps, many algorithms still generated images containing cats (Fig.~\ref{fig:qual_eval}), indicating incomplete unlearning of this larger category. For methods like SA that require specifying a target distribution for forgotten prompts, we use a gray background to increase the difficulty of the unlearning task.

These experiments are relatively lightweight, utilizing a batch size of 32 and leveraging model parallelism across 8 TPU-v3 cores.  We employ deterministic training and maintain consistent data order across all algorithms for each experimental run to further enhance reproducibility.

\paragraph{Evaluation Metrics}

We assess the quality of the unlearned models using FID@30k and CLIPScore. To measure the divergence from the original pre-trained model, we utilize $\consmet$.  To quantify unlearning accuracy, we employ a classifier-based approach, as used in prior work~\citep{fan2024salun, Zhang2024UnlearnCanvasAS}. Instead of reporting Unlearning Accuracy ($UA$), we report $\unlmet = 1 - UA$. This provides a more informative measure, especially when $\unlmet$ is very close to zero. The $\unlmet$ metric is computed using task-specific classifiers:

\begin{itemize}
    \item Animal: We fine-tune a ViT Huge model~\citep{vit2021} on ImageNet to detect the presence of ``cat''.
    \item We fine-tune VGG16~\citep{vgg2014} on images generated in various artistic styles to identify the presence of Van Gogh's style.
    \item We adopt the triplet-loss approach from VGG-Face~\citep{vggface2015}, comparing generated images against a reference image of the target celebrity.
\end{itemize}

For each task, we evaluate unlearning performance on a set of 128 prompts from $\dataf$ that, when used with the baseline model, generate images containing the concept to be forgotten. While quantitative metrics like $\unlmet$ provide valuable insights, they may not fully capture all aspects of unlearning. We discuss these limitations and the need for more comprehensive evaluation metrics in \S\ref{par:qualitative}, with the aim of addressing these challenges in future work.

\paragraph{Quantitative Comparison}
Table~\ref{tab:quant_comp} presents a quantitative comparison across methods.  Regarding image quality, we aim to maintain an FID score close to that of the original checkpoint. EraseDiff can potentially improve FID due to the sampling process from $\datar$. Both \mixeralgo and \overwritealgo successfully maintain FID scores close to the original checkpoint on the Celebrity and Artist tasks. As a general guideline, FID differences within $[0,0.5]$ are typically imperceptible, while those exceeding $1.0$ are noticeable.  For $\consmet$, we estimated variance using 3 runs and found it to be upper-bounded by $0.02$, indicating that our methods yield statistically significant improvements over the baselines. NegGrad, ESD, and SalUn often produce unacceptable results in terms of FID. On the Animal task, NegGrad, ESD, SalUn, and EraseDiff fail to effectively unlearn the cat category (unlearning progress stalled, as indicated by a lack of significant changes in generated images corresponding to cat images as the
method converged), see Figure~\ref{fig:qual_eval}.  Conversely, methods that explicitly define the target distribution (SA, \overwritealgo) generally achieve higher $\unlmet$ values across tasks but require a larger number of steps
($\convsteps$) to converge.

\numparagraph{Qualitative Considerations}\label{par:qualitative}
While $\unlmet$ provides a quantitative measure of unlearning, it does not fully capture the qualitative aspects of the generated images, as demonstrated in Figures~\ref{fig:qual_eval} and \ref{fig:fid_inadequate}. For instance, despite achieving very low $\unlmet$ values with EraseDiff, residual features of the cat and Van Gogh's painting from the original checkpoint remain discernible. Furthermore, unsupervised methods like \mixeralgo lack the ability to specify a target distribution. This leads to unpredictable outcomes, such as generating a sketch for Van Gogh and a ``blob of fur'' for the cat.  In practical applications, methods that allow for explicit control over the target distribution may be preferable, even though they require more steps to converge ($\convsteps$ in Table~\ref{tab:quant_comp}).

\numparagraph{Ablation: Lack of Access to pre-training Data}\label{par:abl_pre_data}
We investigate the scenario where the original pre-training images ($\datar$) are unavailable, and an approximate retain dataset ($\datargen$) is generated from diverse prompts using the base checkpoint
$\theta_0$. This approach, employed in previous work (SA, \citep{heng2023selective}), addresses situations where access to the original training data is restricted, such as with Stable Diffusion.
While our primary experiments in Table~\ref{tab:quant_comp} assume access to $\datar$, it is crucial to assess the impact on FID scores when using $\datargen$. A common approach, as seen in~\citep{heng2023selective}, is to utilize the standard diffusion loss ($\diffloss$) on $\datargen$. In contrast, we employ our proposed consistency loss ($\consloss$).  Our findings demonstrate that $\consloss$ leads to significant improvements in FID scores: a reduction of $3.8$ for Celebrity, $1.4$ for Artist, and $3.9$ for Animal. These gains are statistically significant, with variance across 3 runs upper-bounded by $0.2$ in each case.

\numparagraph{Ablation: Impact of Removing Help Data}\label{par:abl_help}
We analyze the effect of removing the $\helploss$ term from our framework. This ablation reveals interference with the retain dataset ($\datar$), as illustrated in Figure~\ref{fig:fid_inadequate}. While the content of Vermeer's painting is preserved, the image is rendered in grayscale, indicating interference.

\numparagraph{Construction of $\prompth$}\label{par:construction_help_prompts}
For each task, we generated 1024 candidate prompts using Gemini Pro (API accessed between April and July 2024). We employed a simple rule-based approach to generate these prompts. For example, in the Animal task, we used prompts featuring other animal species (except ``frogs''), and in the Artist task, we used prompts referencing artistic works from painters other than Vermeer (the exclusion of frogs and Vermeer was to avoid data overlap for Figures~\ref{fig:fid_inadequate} and \ref{fig:qual_eval}).
We then embedded these candidate prompts using SentenceT5 XXL~\citep{ni-etal-2022-sentence-t5}, a different model from the T5 text encoder used in our main framework, and selected the 256 prompts closest in the embedding space to those in $\promptf$.
\section{Conclusion}
\label{sec:conclusion}

Despite their initial promise, our findings demonstrate that approximate Machine Unlearning approaches struggle to maintain model integrity. Moreover, we have highlighted that evaluating retention solely through FID and CLIPScore can be misleading. To address this, we introduced a new metric, $\consmet$, that directly measures the perceptual distance between images generated by the unlearned model and the original checkpoint. Ideally, effective unlearning methods would yield a low $\consmet$ distance on the retain set $\datar$, indicating preserved model integrity, alongside a low probability of generating undesired concepts, represented by $\unlmet$. Our proposed unlearning methods, \mixeralgo and \overwritealgo, outperform existing baselines in terms of both retention and unlearning efficacy, across both unsupervised and supervised settings. However, we also acknowledge the challenges of unlearning categories associated with larger  more pervasive datasets, such as ``cat''. Current methods appear to struggle in achieving a satisfactory balance between unlearning and retention in such cases. This underscores the need for further research and development in this field. While our work highlights the current limitations of approximate Machine Unlearning, it also provides a foundation for future advancements. The introduction of our novel metric and the demonstration of improved unlearning methods offer promising avenues for enhancing model integrity while effectively removing undesired concepts.
\section{Limitations}
In this study, we deliberately focused on the scenario where the original network remains unmodified, and the goal is to remove concepts solely from the original checkpoint. Furthermore, we restricted our exploration to the image decoder, excluding techniques like text-inversion attacks, adversarial losses, and concept editing that target the text encoder. This limitation aligns with real-world situations where the text encoder might be a large, pre-trained language model that is kept frozen, particularly if the original text embeddings have already been pre-computed on the pre-training data. Additionally, we acknowledge that the unlearning metric $\unlmet$, while effective in quantifying the probability of generating the undesired concept and rooted in previous work, does not account for more nuanced qualitative features. Therefore, we advocate for further research to develop a comprehensive evaluation framework that can quantitatively assess the qualitative aspects of images generated after unlearning, especially in response to prompts from the forget set.
\section{Acknowledgements}
We thank Katja Filippova and Eleni Triantafillou for thoughtful feedback on our draft.
\section{Contributions}
Andrea Schioppa implemented the new algorithms, baselines and ran the experiments. Andrea Schioppa, Emiel Hoogeboom and Jonathan Heek developed the integrity metric and the new unlearning algorithms.

\bibliography{main}
%%%%%%%%%%%%%%%%%%%%%%%%%%%%%%%%%%%%%%%%%%%%%%%%%%%%%%%%%%%%%%%%%%%%%%%%%%%%%%%
%%%%%%%%%%%%%%%%%%%%%%%%%%%%%%%%%%%%%%%%%%%%%%%%%%%%%%%%%%%%%%%%%%%%%%%%%%%%%%%
% APPENDIX
%%%%%%%%%%%%%%%%%%%%%%%%%%%%%%%%%%%%%%%%%%%%%%%%%%%%%%%%%%%%%%%%%%%%%%%%%%%%%%%
%%%%%%%%%%%%%%%%%%%%%%%%%%%%%%%%%%%%%%%%%%%%%%%%%%%%%%%%%%%%%%%%%%%%%%%%%%%%%%%
\newpage
\appendix
% Makes a TOC for the Appendix
\addtocontents{toc}{\protect\setcounter{tocdepth}{2}}
\tableofcontents
\section{Appendix}

\subsection{Experiment Hyper-parameters}
\label{sec:algo_hyperparameters}
We report our experiment hyper-parameters. We used AdamW optimizer using the default values in the optax implementation.
The learning rate was set to $5\times 10^{-7}$ except for ESD for which we used $2\times 10^{-5}$. The global batch size
was 32 and we used both model parallelism (4-way) and parameter sharding (see~\cite{Hoogeboom2023simplediffusion} for details).
For $\beta$ we did preliminary experiments and then fixed a value of $\beta=10$ throughout the experiments.

\subsection{Algorithm implementation}
\label{sec:algo_implementation}
We describe our implementation in Jax.

First, we rely on the following modules.

\begin{python}
import dataclasses
import jax
import jax.numpy as jnp
from typing import Dict, Protocol, Sequence, Tuple
import functools
import flax
\end{python}

This is the implementation of the integrity metric.

\begin{python}
prompts: Sequence[str]
seeds: Sequence[str]

def generator(prompt: str, seed: str, theta) -> jnp.ndarray:
  """Generates an image using checkpoint theta."""
  pass

def lpips_scorer(img_1: jnp.ndarray, img_2: jnp.ndarray) -> jnp.ndarray:
  """Returns the LPIPS distance between img_1 and img_2."""
  pass

def compute_integrity(theta_0, theta_u) -> float:
  """Computes I between theta_u wrt. to theta_0"""

  lpips_scores = []
  for p, s in zip(prompts, seeds):
    sample_ref = generator(p, s, theta_0)
    sample_unl = generator(p, s, theta_u)
    lpips_scores.append(lpips_scorer(sample_ref, sample_unl))
  lpips_scores = jnp.concatenate(lpips_scores, axis=0)
  return jnp.mean(lpips_scores)
\end{python}

Here is some boiler plate that abstract the diffusion model and the training
losses.

\begin{python}

@flax.struct.dataclass
class ModelOutput():
  eps: jnp.ndarray  # noise eps
  eps_theta: jnp.ndarray  # model noise prediction

class ModelFn(Protocol):
  def __call__(
    self, batch: Dict[str, jnp.ndarray], theta: jnp.ndarray,
    rng: jax.random.PRNGKey) -> ModelOutput:
    """Represents applying the model."""
    pass

def optimizer_fn(grad, theta, learning_rate, step, optimizer_state):
  """Updates the parameters and the optimizer state using the gradient."""
  pass

def diff_loss(
  theta, # Model params
  rng: jax.random.PRNGKey,
  model_fn: ModelFn,
  batch: Dict[str, jnp.ndarray],
  ) -> jnp.ndarray:
  """L_diff loss in the paper."""
  model_output = model_fn(batch, theta, rng)
  eps = model_output.eps
  eps_theta = model_output.eps_theta
  return jnp.mean(jnp.square(eps - eps_theta))

def integrity_loss(
  theta, # Model params
  model_fn: ModelFn,
  rng: jax.random.PRNGKey,
  batch: Dict[str, jnp.ndarray],
  theta_0, # Base checkpoint parameters
):
  """L_I loss in the paper."""

  eps_theta_u = model_fn(
      batch, theta_u, rng).eps_theta
  eps_theta_0 = model_fn(
      batch, theta_0, rng).eps_theta

  return jnp.mean(jnp.square(eps_theta_u - eps_theta_0))
\end{python}

We then describe the implementation of \mixeralgo. We start with the configuration:
\begin{python}
@dataclasses.dataclass
class SaddleConfig:
  """Saddle algorithm config."""

  learning_rate: float
  beta: float = dataclasses.field(
      metadata=dict(help='Strength of the integrity term.')
  )
\end{python}

We then define the training step:
\begin{python}
def saddle_step(
  config: SaddleConfig,
  model_fn: ModelFn,
  base_rng: jax.random.PRNGKey,
  theta, # Model parameters
  theta_0, # Base checkpoint parameters
  optimizer_state, # state of the optimizer
  step: jnp.ndarray, # Training step for optimizer schedule
  retain_batch: Dict[str, jnp.ndarray],
  forget_batch: Dict[str, jnp.ndarray],
):
  """Saddle implementation."""

  rng = jax.random.fold_in(base_rng, jax.lax.axis_index('batch'))
  rng = jax.random.fold_in(rng, step)

  # Loss and gradient
  rng_keep, rng_forget = rng.split(2)

  retain_loss_fn = functools.partial(
      integrity_loss, model_fn=model_fn,
      rng=rng_keep, batch=retain_batch, theta_0=theta_0)
  forget_loss_fn = functools.partial(
      diff_loss, model_fn=model_fn,
      rng=rng_forget, batch=forget_batch)

  grad_retain = jax.grad(retain_loss_fn)(theta)
  grad_forget = jax.grad(forget_loss_fn)(theta)
  # Note the change of sign to grad_forget to do ascent.
  grad = jax.tree_util.tree_map(
      lambda x, y: config.beta * x - y, grad_retain, grad_forget
  )

  new_theta, new_optimizer_state = optimizer_fn(
    grad=grad_clip , theta=theta, learning_rate=config.learning_rate, step=step,
    optimizer_state=optimizer_state)

  return new_theta, new_optimizer_state
\end{python}

We then describe the implementation of \overwritealgo. We start with the configuration:

\begin{python}
@dataclasses.dataclass
class OVWConfig:
  """OVW algorithm config."""

  learning_rate: float
  beta: float = dataclasses.field(
      metadata=dict(help='Strength of the integrity term.')
  )
\end{python}

We then define the training step (which takes two steps instead of one):

\begin{python}
def ovw_step(
  config: OVWConfig,
  model_fn: ModelFn,
  base_rng: jax.random.PRNGKey,
  theta, # Model parameters
  theta_0, # Base checkpoint parameters
  optimizer_state, # state of the optimizer
  step: jnp.ndarray, # Training step for optimizer schedule
  retain_batch: Dict[str, jnp.ndarray],
  overwrite_batch_1: Dict[str, jnp.ndarray],
  overwrite_batch_2: Dict[str, jnp.ndarray],
  help_batch: Dict[str, jnp.ndarray],
):
  """OVW implementation.

  The overwrite_batch is the forget_batch with the target modified to
  target the desired distribution.

  The retain_batch is sampled from the retain dataset.
  The help_batch is sampled from the help dataset D_h.

  Note that this function takes two steps.
  """

  rng = jax.random.fold_in(base_rng, jax.lax.axis_index('batch'))
  rng = jax.random.fold_in(rng, step)

  # Loss and gradient
  rng_retain, rng_overwrite_1, rng_overwrite_2, rng_help = rng.split(4)

  grad_retain = jax.grad(integrity_loss)(
      theta, batch=retain_batch, rng=rng_retain, model_fn=model_fn,
      theta_0=theta_0)
  grad_overwrite_1 = jax.grad(diff_loss)(
      theta, batch=overwrite_batch_1, rng=rng_overwrite_1, model_fn=model_fn)
  grad = jax.tree_util.tree_map(
      lambda x, y: config.beta * x + y, grad_retain, grad_overwrite_1
  )
  theta, optimizer_state = optimizer_fn(
    grad=grad_clip , theta=theta, learning_rate=config.learning_rate, step=step,
    optimizer_state=optimizer_state)

  grad_help = jax.grad(integrity_loss)(
      theta, batch=help_batch, rng=rng_retain, model_fn=model_fn,
      theta_0=theta_0)
  grad_overwrite_2 = jax.grad(diff_loss)(
      theta, batch=overwrite_batch_2, rng=rng_overwrite_2, model_fn=model_fn)
  grad = jax.tree_util.tree_map(
      lambda x, y: config.beta * x + y, grad_help, grad_overwrite_2
  )

  new_theta, new_optimizer_state = optimizer_fn(
    grad=grad_clip , theta=theta, learning_rate=config.learning_rate,
    step=step+1,
    optimizer_state=optimizer_state)

  return new_theta, new_optimizer_state
\end{python}

\end{document}